\documentclass[10pt,twocolumn,letterpaper]{article}

\usepackage[pagenumbers]{cvpr} 

\usepackage{epsfig}
\usepackage{graphicx}
\usepackage{amsmath}
\usepackage{amssymb}
\usepackage{bm}
\usepackage{url}
\usepackage{amsfonts}
\usepackage{amstext}
\usepackage{multirow}
\usepackage{microtype}
\usepackage{booktabs}
\usepackage{pifont}
\usepackage{array}
\usepackage{indentfirst}
\usepackage[american]{babel}
\usepackage[ruled, linesnumbered]{algorithm2e}

\usepackage[pagebackref,breaklinks,colorlinks]{hyperref}

\usepackage[capitalize]{cleveref}
\crefname{section}{Sec.}{Secs.}
\Crefname{section}{Section}{Sections}
\Crefname{table}{Table}{Tables}
\crefname{table}{Tab.}{Tabs.}

\def\cvprPaperID{0000} 
\def\confName{CVPR}
\def\confYear{2023}

\begin{document}

\title{Dynamic Token-Pass Transformers for Semantic Segmentation}

\author{
   Yuang Liu\textsuperscript{\rm 1}\thanks{The work was done as a Research Intern at DAMO Academy, Alibaba.} \,, Qiang Zhou\textsuperscript{\rm 2}, Jing Wang\textsuperscript{\rm 2}, Zhibin Wang\textsuperscript{\rm 2}, Fan Wang\textsuperscript{\rm 2}, Jun Wang\textsuperscript{\rm 1}, Wei Zhang\textsuperscript{\rm 1}\thanks{Corresponding author.} \\
\textsuperscript{\rm 1}East China Normal University \ \textsuperscript{\rm 2}DAMO Academy, Alibaba Group \\
\texttt{\{frankliu624,zhangwei.thu2011,wongjun\}@gmail.com} \\
\texttt{\{jianchong.zq,yunfei.wj,zhibin.waz,fan.w\}@alibaba-inc.com} \\
}

\maketitle

\begin{abstract}
   Vision transformers (ViT) usually extract features via forwarding all the tokens in the self-attention layers from top to toe. In this paper, we introduce dynamic token-pass vision transformers (DoViT) for semantic segmentation, which can adaptively reduce the inference cost for images with different complexity. DoViT gradually stops partial easy tokens from self-attention calculation and keeps the hard tokens forwarding until meeting the stopping criteria. We employ lightweight auxiliary heads to make the token-pass decision and divide the tokens into keeping/stopping parts. With a token separate calculation, the self-attention layers are speeded up with sparse tokens and still work friendly with hardware. A token reconstruction module is built to collect and reset the grouped tokens to their original position in the sequence, which is necessary to predict correct semantic masks. We conduct extensive experiments on two common semantic segmentation tasks, and demonstrate that our method greatly reduces about 40\% $\sim$ 60\% FLOPs and the drop of mIoU is within 0.8\% for various segmentation transformers. The throughput and inference speed of ViT-L/B are increased to more than 2$\times$ on Cityscapes. 
\end{abstract}

\section{Introduction}

Semantic segmentation has been a significant component of autonomous driving~\cite{levinson2011towards}, image editing~\cite{zhang2020image} and visual scene analysis~\cite{chen2019towards}. As a dense prediction task, it aims to assign each image pixel to a category label. Thanks to the development of deep neural networks, especially vision transformers (ViT)~\cite{dosovitskiy2020image}, the research for semantic segmentation has achieved great successes at the price of huge computation. Moreover, the transformer-like segmentor, \eg, SETR~\cite{zheng2021rethinking}, Segmenter~\cite{strudel2021segmenter} and Segformer~\cite{xie2021segformer}, has overtaken CNN across the board and shows great potential. However, the computational complexity of the transformer architecture makes real-time application of semantic segmentation even more prohibitive. To make these models more suitable to resource-constrained mobile devices, it is urgent to reduce the computation cost and accelerate them. 

\begin{figure}[!t]
   \centering
   \subfloat[Parameter-aware]{
      \includegraphics[width=0.42\linewidth]{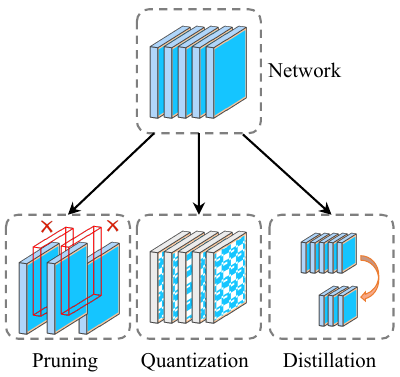}
      \label{fig:intro_a}
   }
   \subfloat[Data-aware (Ours)]{
      \includegraphics[width=0.52\linewidth]{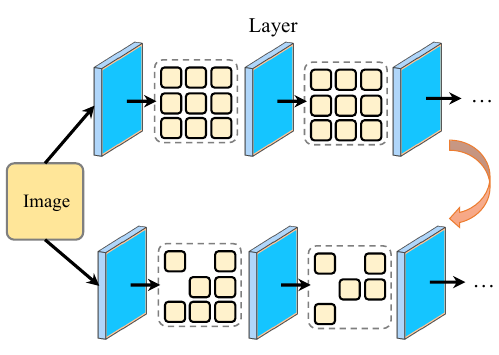}
      \label{fig:intro_b}
   }
   \caption{Overview of different algorithms for model acceleration.}
   \label{fig:intro}
\end{figure}

These years have witnessed the great progress in CNN-type model compression and acceleration brought by parameter-aware approaches. As shown in Figure~\ref{fig:intro_a}, the majority of current acceleration approaches can be divided into three categories, including pruning~\cite{li2017pruning,lin2020hrank}, quantization~\cite{wang2019haq,jin2020adabits} and knowledge distillation~\cite{hinton2015distilling,gou2021knowledge,wang2020knowledge}. They all focus on reducing the redundant components or parameters of the networks, so we called them parameter-aware acceleration methods. There has been a recent surge of interest to introduce these parameter-aware acceleration methods to transformer-base architectures, both in natural language processing (NLP)~\cite{shen2020qbert,liu2020fastbert} and computer vision (CV)~\cite{liu2021post,yu2021unified}.

The core of ViTs is the self-attention module, which is naturally different from the convolution operation in CNNs. It works by calculating the relationships among each pair of image patches or tokens, and then capturing the global context of the input image. Benefiting from this nature of self-attention, Rao~\etal firstly propose DynamicViT~\cite{rao2021dynamicvit} that prunes the tokens of less importance and only keeps partial tokens in self-attention for acceleration. A-ViT~\cite{yin2022avit} improves DynamicViT~\cite{rao2021dynamicvit} by introducing halting distribution of tokens and requires no extra parameters. Recently, ATS~\cite{fayyaz2021ats} builds an adaptive token sampler to automatically select the most important tokens. These data-aware acceleration works point out a new direction for model acceleration. But they only pay attention to classification tasks, and do not support dense tasks like semantic segmentation. For classification ViT, actually only the class token is utilized to predict the category of the whole image, and most of the other tokens may be dropped at certain layers. Differently, each token is necessary to semantic segmentation, and all the tokens are required to be utilized by the decoder to predict the categories of pixels. Additionally, the above methods train the dynamic ViT with a fixed token pruning rate for each layer/block, making it unable to make an image-wise trade-off between the input complexity and inference cost. 

To this end, we propose a novel dynamic token-pass transformers (DoViT) for semantic segmentation. This is the first attempt of data-aware ViT acceleration on dense prediction tasks. Rather than focusing on the patch redundancy in the input image, we base complexity or learning difficulty of semantic patches/tokens to adaptively determine their computational cost, which makes the backbone achieve an image-wise dynamic inference. 
It's difficult to automatically select the more informative or important tokens for self-attention at each layer. Because the pixel classification may fail if any semantic token is not fully learned. The most reliable scheme is to decide whether a token should be preserved or stopped explicitly based on the early prediction results. Therefore, we introduce a semantic early-probe scheme that divides the tokens into two categories, \ie, keeping set and stopping set. The keeping tokens will involve in the next self-attention layers while the stopping tokens will be prevented from the self-attention and directly passed to the decoder via a short path. The keep or stop of each token is determined by the semantic prediction confidence provided by the auxiliary heads. This scheme makes it possible to adjust a fully dynamic inference cost for various input images. 
We claim that all semantic tokens must be received by the decoder in their original position after a certain level of self-attention operations. A separate self-attention module is presented to optimize the calculation of the keeping/stopping set of tokens where the stopping tokens bring no computation. To restore the order of input tokens, we introduce a token reconstruction module that provides complete and correct feature maps for the decoder and auxiliary heads. 
Our proposed approach significantly cuts down the inference cost --- the FLOPs of SETR on Cityscapes is reduced by 40\% $\sim$ 60\% within 0.8\% mIoU drop, and the throughput and FPS is improved to over 2 $\times$ on hardware. 
In summary, our main contributions are as follows:
\begin{itemize}
   \item We propose a dynamic token-pass method to reduce the inference cost of vision transformers for semantic segmentation. 
   \item We introduce a semantic early-probe scheme to determine the token-pass candidates. The separate self-attention and token reconstruction modules are responsible for sparse token acceleration.
   \item We conduct extensive experiments on two public segmentation datasets with various ViT models and demonstrate that the proposed method reduces FLOPs significantly with minor drop in mIoU. 
\end{itemize}

\section{Related Work}

\subsection{Semantic Segmentation}

Fast development of deep neural network has significantly inspired the exploration of semantic image segmentation. At the very first, FCN~\cite{long2015fully} achieves pixel-wise image segmentation by removing the final fully-connected layer. As FCN focuses on extracting abstract semantic features, multi-scale feature fusing~\cite{badrinarayanan2015deep, noh2015learning, ronneberger2015u}, dilated convolution~\cite{chen2017deeplab, chen2014semantic, yu2015multi}, and spatial pyramid pooling ~\cite{chen2017rethinking, chen2018encoder, zhao2017pyramid} are proposed to induce the network to extract more fine-grained features. To further boost the accuracy of the network, attention mechanisms are introduced to the network~\cite{fu2020scene, fu2019dual, yin2020disentangled, yu2020context, yuan2018ocnet, zhao2018psanet}. 
Recently, considering the excellent performance of transformers, many works try to plug it into the semantic segmentation.
SETR~\cite{zheng2021rethinking} firstly adopts a transformer based network, ViT, as encoder to extract features, but keep the CNN-based decoder. The Segmenter~\cite{strudel2021segmenter} further expands the transformer architecture to the decoder and designs a pure transformer encoder-decoder segmentation architecture. 

\subsection{Model Acceleration}

Even transformer has brought great improvement for the semantic segmentation, the quadratic number of interactions between tokens increased the computation burdens. To promote the deployment of transformer-based model on edge devices, model acceleration become a popular topic. Traditional parameter-aware model acceleration methods like knowledge distillation~\cite{hinton2015distilling,wang2020knowledge,gou2021knowledge}, quantization~\cite{wang2019haq,jin2020adabits}, and pruning~\cite{li2017pruning,he2018soft,lin2020hrank} have already been introduced to transformer~\cite{shen2020qbert,liu2020fastbert,liu2021post,yu2021unified,meng2022adavit}. Considering the cost of calculating relations among image patches or tokens in transformer, data-aware model acceleration is worth being discussed. Some related works have been studied in image classification, for example, DynamicViT~\cite{rao2021dynamicvit} firstly proposes to prune uninformative tokens in a dynamic way by adopting a lightweight prediction module. Then A-ViT~\cite{yin2022avit} improves it by removing the prediction module and halting the computation of tokens by a parameter-free adaptively inference mechanism. EViT~\cite{liang2022evit} reduces computation by progressively discarding or fusing inattentive tokens in Vision Transformers. Recently, ATS~\cite{fayyaz2021ats} builds an adaptive token sampler to automatically select the most important tokens. These methods exhibit great potential in transformer acceleration of classification tasks, but they are not suitable for the semantic segmentation as each token is meaningful for the pixel-wise prediction.

\begin{figure*}[!t]
   \centering
   \includegraphics[width=\linewidth]{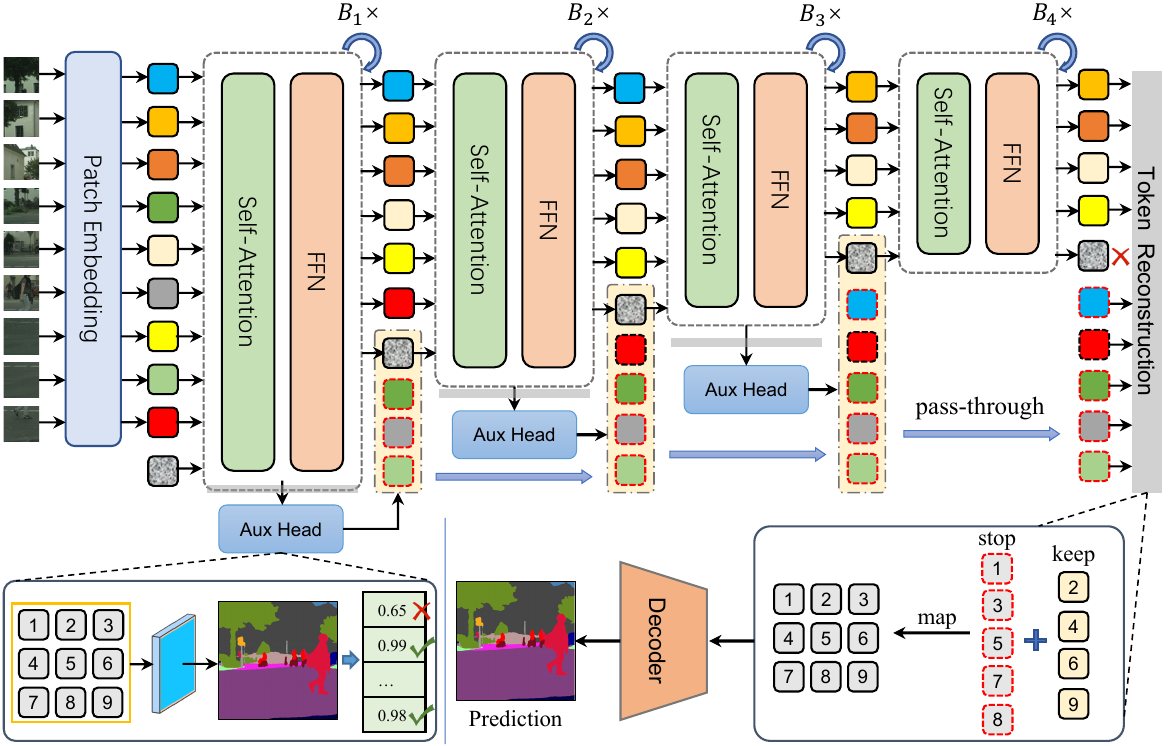}
   \caption{The overview of our dynamic token-pass transformers for semantic segmentation. The token-pass decision depends on the auxiliary (Aux) heads inserted between the transformer blocks. With the keeping and stopping token sets, the separate self-attention module can capture the more informative context with less computation, which represents the core of our dynamic token-pass algorithm. The token reconstruction module is responsible for converting the sparse tokens sets to a structured feature map with ordered tokens.}
   \label{fig:framework}
\end{figure*}

\section{Proposed Approach}

\subsection{Overview}

Figure~\ref{fig:framework} illustrates the pipeline of our DoViT framework. Consider a transformer-like segmentation network that takes an image $x \in \mathbb{R}^{3\times H\times W}$ as input to predict a semantic mask $y \in \mathbb{R}^{C\times H\times W}$ ($C$, $H$, and $W$ represent number of categories, height, and width respectively): 
\begin{equation}
   y = \mathcal{D} \odot \mathcal{F}^{L} \odot \mathcal{F} ^{L-1} \odot \mathcal{F}^{1} \odot \mathcal{E}(x) \,,
\end{equation}
where $\mathcal{E}, \mathcal{F}$ and $\mathcal{D}$ are the patch embedding network, backbone layer and segmentation decoder, respectively, $L$ represents the number of layers. The network $\mathcal{E}$ tokenizes the image patches from $x$ into positioned tokens $T\in\mathbb{R}^{N\times E}$, where $N$ and $E$ are the number and dimension of the tokens, respectively. Then the tokens forward with multi-head self-attention (MSA) operations in the backbone layers, which accounts for most of the computation of the entire segmentation network. We select certain layers and execute a semantic early-probe scheme to make a token-pass decision that the hard tokens are kept in the later MSA layers while the easy tokens are stopped from MSA and directly passed to the decoder. To perform an efficient sparse token forwarding in the backbone and decoder network, we introduce a separate self-attention strategy and a token reconstruction module.

\subsection{Token-Pass Decision}

To progressively lessen the tokens in the ViT backbone, we choose $D$ nonadjacent self-attention layers as ``decision layers'' that divide the whole $L$ backbone layers into $D+1$ blocks, \ie, $\{ \mathcal{B}^1, \mathcal{B}^2, \cdots, \mathcal{B}^{D+1} \}$. The number of layers in the block $\mathcal{B}^\ell$ is $B_\ell$. 
Different from classification ViT, it's a challenge to determine the importance of each token at certain layers. Because every token contains specific semantic information and contributes to the segmentation prediction, which indicates that it's unreasonable to roughly drop the uninformative tokens. We claim that it is crucial to accurately judge whether a token is fully utilized and learned. A straightforward approach is to evaluate the segmentation results with the early tokens. To this end, we propose an early-probe scheme to determine whether a token is kept in calculating or stopped learning. In particular, we insert a lightweight auxiliary segmentation head $\mathcal{H}^\ell$ following the block $\mathcal{B}^\ell$. It's worth noting that the auxiliary head consisting of one fully-convolutional layer brings tiny extra parameters and computation, which is almost negligible comparing to the whole segmentation network. 
Assuming that no token-pass decision is applied, and each MSA block works normally with total $N$ tokens input and $N$ tokens output. We reshape the token sequence $T^{\ell}$ output by the block $\mathcal{B}^{\ell}$ to a deep feature map $\bm{f} \in \mathbb{R}^{E\times h \times w}$, where $E$ is both the dimension of the tokens and channels of the feature map, $h$ and $w$ represents the width and height of the feature map, $N = hw$. 
We feed the feature map $\bm{f}^\ell$ into the $\ell$-th auxiliary head $\mathcal{H}^\ell$ to obtain a probability map $p^\ell \in \mathbb{R}^{C \times h \times w}$ with the softmax operation:
\begin{equation}
   p^\ell = \text{softmax}\left(\mathcal{H}^{\ell}(\bm{f}^\ell)\right) \,.
\end{equation}
The scalar $p_{c,i,j}$ represents the probability of the token $T_{i,j}$ belonging to the $c$-th  semantic category, and the maximum value $q_{i,j}$ of $\{p_{c,i,j}\}_{c=1}^C$ represents the confidence of label prediction in terms of $T_{i,j}$. With these insights, we can allocate a prediction confidence to each token $T_{i,j}$ and obtain a score map 
\begin{equation}
   q^{\ell}_{i,j} = \max \left\{p^{\ell}_{c, i, j} | c \in \{1, 2, \cdots, C\}\right\} \,.
\end{equation}

Generally, the tokens with high prediction confidence are uninformative and easy to segment, while the tokens with low prediction confidence are complex and hard to learn. So we can utilize the confidence score to determine a token's pass. To align with the original shape of token sequence $T\in\mathbb{R}^{N\times E}$, we reshape the confidence map $q^{\ell}\in \mathbb{R}^{h\times w}$ to $q^{\ell}\in \mathbb{R}^{N}$. We define a binary decision mask $\tilde{M}^{\ell} \in \{0, 1\}^N$ to indicate whether to keep or stop each token $T^{\ell}_n$ at the $\ell$-th block, 
\begin{equation}
	\tilde{M}^{\ell}_n = \left\{
		\begin{aligned}
		0 &, \text{if}\,\, q^{\ell}_n > \xi \\ 
		1 &, \text{else} 
	\end{aligned}
	\right. \,,
\end{equation}
where $0 \leq \xi \leq 1$ is a threshold parameter. 
The number of sparsely-keeping tokens is gradually decreased block-by-block. But the decision mask $\tilde{M}^{\ell}$ is calculated based on the hypothesis that the head $\mathcal{H}^\ell$ receives a complete feature map without token reduction. So we need to ignore the tokens that are stopped at the previous blocks by updating $\tilde{M}^{\ell}$ with
\begin{equation}
   M^{\ell} = \tilde{M}^{\ell} \odot M^{\ell -1} = \prod_{i=0}^{\ell} \tilde{M}^{i} \,, 
\end{equation}
in which $\odot$ is the Hadamard product, and $M^{\ell -1}$ is the real decision mask from the last block, $M^0 = \tilde{M}^0 = \bm{1}$.  

In this way, we have selected the keeping tokens at each block, they will involve in the MSA blocks until meeting the stopping criteria. With the early-probe scheme, the token-pass is dynamic adaptively in terms of the patch complexity, rather than subject to a fixed keeping/stopping ratio at each stage. It's necessary to collect the stopping tokens to build a complete feature map for the auxiliary heads and segmentation decoder. To achieve an efficient self-attention with the keeping/stopping tokens, we design a separate token forwarding algorithm.

\subsection{Sparse Token Forwarding}

With the decision masks $\{\tilde{M}^\ell\}^D_{\ell}$, a sparse token sequence can be extracted from the blocks. \cite{rao2021dynamicvit} and \cite{yin2022avit} both execute the self-attention operation with the sparse tokens via a mask mechanism, where the exiting/stopping tokens still involve in the calculations of query, key and value. To achieve a real token reduction, they design different training and inference phases in which the stopping tokens are inconsistent. The inconsistent tokens input to the decoder could make a bias between the training and inference phases. To this end, we propose a simple yet efficient separate self-attention module that keeps the consistency of dynamic token-pass in two phases. 

\textbf{Separate Self-Attention.} 
The keeping and stopping tokens are processed separately in each block, except the first block $\mathcal{B}^1$ in which no token reduction is performed. 
To gather the sparse tokens to a compact and structured representation, we define a function $\mathcal{G}(T, M)$ that selects tokens from $T \in \mathbb{R}^{N\times E}$ with the mask $M$ and combine them to a new token sequence with the size of $|M| \times E$, where $|M|$ is the number of nonzero items. The keeping/stopping token sequence $\hat{T}^{\ell}/\ddot{T}^{\ell}$ up to the $\ell$-th block can be obtained by 
\begin{equation}
   \hat{T}^{\ell} = \mathcal{G}\left(T^\ell, M^\ell\right) \,, \, \ddot{T}^{\ell} = \mathcal{G}\left(T^\ell, (\bm{1}-M^\ell)\right) \,.
\end{equation}
In this way, the keeping/stopping tokens output by the block $\mathcal{B}^\ell$ has been divided into two parts. 
In dynamic token-pass inference, only the keeping tokens are considered in the MSA modules of the next block, \ie $\mathcal{B}^{\ell+1}$, which can be formulated as 
\begin{equation}
   \text{MSA}(\hat{\mathbf{Q}}, \hat{\mathbf{K}}, \hat{\mathbf{V}}) = \text{softmax}\left(\frac{\hat{\mathbf{Q}}\hat{\mathbf{K}}^\top}{\sqrt{d}} \right)\hat{\mathbf{V}} \,,
\end{equation}
where $\hat{\mathbf{Q}}, \hat{\mathbf{K}}, \hat{\mathbf{V}} \in \mathbb{R}^{N\times d}$ are the query, key and value embeddings of the keep tokens $\hat{T}$, $d$ is the dimension of the embeddings. Then, the computational complexity of self-attention is reduced from $\mathcal{O}(N^2\cdot d)$ to $\mathcal{O}(|M|^2\cdot d)$, $|M| \leq N$. 
The stopping tokens $\ddot{T}^\ell$ will directly pass the next block $\mathcal{B}^{\ell+1}$ without any calculation. 
After separate self-attention in block $\mathcal{B}^{\ell+1}$, the keeping and stopping tokens can be combined as a complete sequence $\hat{T}^{\ell+1}$. 

\textbf{Token Reconstruction.}
However, due to the token sparsifcation and separation for self-attention, the combined token sequence $\hat{T}^{\ell+1}$ output by block $\mathcal{B}^{\ell+1}$ is out-of-order and inconsistent with the original image patches. To this end, we introduce a token reconstruction module to locate the tokens in $\hat{T}^{\ell+1}$ to their original position as a new sequence. It can be formulated as $T^{\ell+1} = \mathcal{R}(\hat{T}^{\ell+1}, I^{\ell+1})$, in which $\mathcal{R}$ is a transform function that maps each token to the corresponding position or rank in the sequence, $I^{\ell+1}$ is the map of token indices that can be generated incidentally by $\mathcal{G}$. Finally, the reconstructed tokens $T^{\ell+1}$ can be fed into the next block. After reshaped as a feature map $\bm{f}^{\ell+1}$, the auxiliary head $\mathcal{H}^{\ell+1}$ can utilize it to make token-pass decision for the next block, and the decoder can finish the semantic prediction with it. As shown in Figure~\ref{fig:framework}, we add a token reconstruction module before the auxiliary heads and segmentation decoder. 

\textbf{Token Merging.}
Considering that there may be some useful information provided by the stopping tokens, we merge them as one representative token and aggregate it with the class token before calculating self-attention in the next block. For example, after obtaining a token sequence $T^{\ell}$ from the block $\mathcal{B}^\ell$ via the above algorithms, the class token can be updated by
\begin{equation}
   T^{\ell}_0 \leftarrow \frac{1}{2} \left(T^{\ell}_0 + \frac{1}{\left\lvert S^{\ell} \right\rvert }\sum^{\lvert S^{\ell} \rvert}_{i} \sum^{E}_{j} \ddot{T}^{\ell}_{i,j} \right) \,,  
\end{equation}
where $|S^\ell| = |\bm{1} - M^\ell|$ is the total number of stopping tokens. Then the keeping tokens including $T^{\ell}_0$ can sequentially involve the self-attention in the next block. 
Note that we preserve the class token in MSA following the standard in ViT~\cite{dosovitskiy2020image}, and remove it when building feature maps. In fact, the class token $T_0$ is a default keeping token in all the blocks, not depends on the token-pass decision. 

\subsection{Training Pipeline}

With the feature maps $\{\bm{f}^\ell\}_{\ell=1}^{D+1}$ consisting of hierarchically-stopping tokens, the auxiliary heads $\{\mathcal{H}^\ell\}^{D}_{\ell=1}$ and segmentation decoder $\mathcal{D}$ predict semantic probability maps $\{p^\ell\}_{\ell=1}^{D}$ and $p^s$.
To train the segmentation network in a supervised manner, the ground truth label map $\bar{y} $ is used to compute the cross-entropy (CE) loss 
\begin{equation}
  \label{eq:ce}
  \mathcal{L}_{CE} (p^s, \bar{y}) = \frac{1}{HW}\sum_{i=1}^{HW} \sum_{j=1}^{C} - \bar{y}_{i,j}\log(p^{s}_{i,j}) \,,
\end{equation}
where $\bar{y}_{i,j}$ is the real value (1 or 0) of the $j$-th class for the $i$-th pixel, and $p^{s}_{i,j}$ corresponds to the probability predicted by the segmentation decoder $\mathcal{D}$. Analogously, the $D$ auxiliary heads are updated with loss function 
\begin{equation}
   \label{eq:ah}
   \mathcal{L}_{AH} = \sum_{\ell}^{D} \mathcal{L}_{CE}(\mathcal{U}(p^\ell), \bar{y}) \,,
\end{equation}
where $\mathcal{U}(\cdot)$ is the upsampling function.

To alleviate the performance damage caused by dynamic token-pass, we employ a self-distillation framework to train the DoViT-based segmentation network with the corresponding ViT-based network as a teacher. We denote the teacher's probability map as $p^t$, then the self-distillation loss is formulated by Kullback-Leibler(KL) divergence: 
\begin{equation}
  \label{eq:sd}
  \mathcal{L}_{SD} (p^s, p^t) = \frac{1}{HW}\sum_{i=1}^{HW}\sum_{j}^{C} p^s_{i,j} \log\left(\frac{p^s_{i,j}}{p^t_{i,j}} \right) \,. 
\end{equation}

The overall loss function is 
\begin{equation}
  \label{eq:okd}
  \mathcal{L} = \mathcal{L}_{CE} + \alpha \mathcal{L}_{AH} + \beta \mathcal{L}_{SD} \,,
\end{equation}
where $\alpha, \beta > 0$ are two hyper-parameters to control the relative importance. 

\section{Experiments}

\subsection{Experimental Settings}

\subsubsection{Datasets and Metrics}

\textbf{Cityscapes}~\cite{cordts2016cityscapes} is a widely used urban scene understanding dataset, with 19 common classes for evaluation. It contains 2,975 fine-annotated images with 1024 $\times$ 2048 pixels for training, 500 for validation, and 1,525 for testing. 

\textbf{ADE20K}~\cite{zhou2017scene} is one of the most challenging semantic segmentation datasets that contains challenging scenes with 20,210 fine-annotated images with 150 semantic classes in the training set. The validation and test set contain 2,000 and 3,352 images respectively. 

\textbf{Metrics.} The numbers of float-point operations (FLOPs) and parameters (Params) are introduced to measure the computational complexity and model size of the segmentation network. We report the throughput and frame-per-second (FPS) to show the inference speed of networks. 
We employ the common metric of mean Intersection over Union (mIoU), Pixel Accuracy (PA), and mean Pixel Accuracy (mPA) for scene segmentation on all datasets.
Note that mIoU is the primary and more persuasive metric for segmentation.

\begin{table*}[!t]
   \centering
   \begin{tabular}{c|l|ccl|l|l}
     \hline
     Network & Backbone & PA (\%) & mPA (\%) & mIoU (\%) & Params (M) $\downarrow$ & FLOPs (G) $\downarrow$ \\
     \hline
     \multirow{9}{*}{SETR} & ViT-L & 95.91 & 85.71 & 78.10 & 305.74 & 2484.27 \\
     & DoViT-L (Ours) & 95.93 & 85.27 & 77.98 (-0.22) & 306.54 (+0.8) & 1088.80 (-56\%) \\
      \cline{2-7}      & ViT-B & 95.65 & 84.22 & 76.59 & 87.62 & 703.28 \\
     & DyViT-B/0.9 & 94.99 & 79.10 & 71.60 (-4.99)  & 91.12 (+3.5) & 626.79 (-11\%) \\
     & DyViT-B/0.85 & 94.82 & 77.44 & 70.11 (-6.48) & 91.12 (+3.5) & 581.84 (-17\%) \\
     & DoViT-B (Ours) & 95.64 & 83.95 & 76.40 (-0.19) & 88.23 (+0.6) & 330.09 (-53\%) \\
      \cline{2-7}      & ViT-S & 95.39 & 83.14 & 74.80 & 22.57 & 177.99 \\
     & DyViT-S/0.9 & 94.10 & 77.52 & 67.47 (-7.33) & 23.62 (+1.1) & 158.87 (-11\%) \\
     & DoViT-S (Ours) & 95.52 & 83.26 & 75.13 (+0.33) & 22.89 (+0.3) & 91.13 (-49\%) \\
     \hline
     \multirow{4}{*}{SETR} & DeiT-B & 95.78 & 85.25 & 77.41 & 87.62 & 703.28 \\
           & DoDeit-B  (Ours) & 95.62 & 84.33 & 76.70 (-0.71) & 88.23 (+0.6) & 332.73 (-53\%) \\
      \cline{2-7}          & DeiT-S & 95.47 & 82.69 & 75.14 & 22.57 & 177.99 \\
           & DoDeit-S  (Ours) & 95.29 & 82.58 & 74.63 (-0.51) & 22.89 (+0.3) & 83.64 (-53\%) \\
     \hline
     \multirow{6}{*}{Segmenter} & ViT-L & 96.07 & 86.41 & 79.10 & 333.82 & 2705.40 \\
           & DoViT-L (Ours) & 95.88 & 85.60  & 78.47 (-0.63) & 334.62 (+0.8) & 1257.86 (-54\%) \\
      \cline{2-7}          & ViT-B & 95.89 & 85.55 & 77.83 & 103.38 & 826.99 \\
           & DoViT-B (Ours) & 95.73 & 84.57 & 77.40 (-0.43) & 103.99 (+0.6) & 442.19 (-47\%) \\
      \cline{2-7}          & ViT-S & 95.68 & 84.22 & 76.61 & 26.47 & 208.05 \\
           & DoViT-S (Ours) & 95.65 & 84.11 & 76.65 (+0.04) & 26.79 (+0.3) & 119.50 (-43\%) \\
     \hline
     \end{tabular}
   \caption{Main results on Cityscapes. Performance comparison of different segmentation
   models with varying backbones on Cityscapes validation set. The ``DoViT'' and ``DoDeiT'' are standard ViT/DeiT backbone with our acceleration method. The suffix ``L/B/S'' represent the large/base/small transformer, respectively.} 
   \label{tab:citys}
\end{table*}

\subsubsection{Implementation Details}

We implement our DoViT frameowrk with PyTorch~\cite{paszke2017automatic} on 8 NVIDIA V100 GPUs. We evaluate the proposed method on two popular transformer-like segmentation architectures, \ie, SETR~\cite{zheng2021rethinking} and Segmenter~\cite{strudel2021segmenter}, using classic ViT~\cite{dosovitskiy2020image} and DeiT~\cite{touvron2021training} as backbone. For the base and small transformer backbone, we select the (3,6,9)-th layers as the decision layers. In the total 24 self-attention layers of the ViT-large, the (6,12,18)-th layers are the decision layers. Three single-layer auxiliary heads with $1\times 1$ kernel are inserted following the decision layers. And the original auxiliary heads for SETR~\cite{zheng2021rethinking} can be reused without specific modification. All the setups of data augmentation, network training and accuracy evaluation follow the offical implementation of SETR~\cite{zheng2021rethinking} and Segmenter~\cite{strudel2021segmenter} in codebase \textit{MMSegmentation}~\cite{mmseg2020}. The co-efficient $\alpha$ and $\beta$ are set to 1.0 and 0.4 by default, respectively. The confidence threshold $\xi=0.985$ is optimal for Cityscapes, and $\xi\in[0.96,0.985]$ is suitable for ADE20K. Without specific instruction, the parameters, FLOPs, throughput and FPS are reported with a $1024\times 2048$ resolution for Cityscapes, and $512\times 512$ randomly cropped images for ADE20K. To test the adaptive inference cost of each image in our DoViT, we randomly sample 100 images from the validation set and report their average FLOPs, throughput and FPS. 
For parallel training of images with various numbers of sparse keeping tokens, we utilize a distributed environment, where the batch size per GPU is set to 1 for Cityscapes and 2 for ADE20K. If the batch size per GPU is larger than 1, the numbers of keeping tokens of batch images on one GPU are set the same, by striking an average according to the sort of confidence.

\subsection{Main Results}

\textbf{Cityscapes.} 
One of the most advantages of our DoViT framework is that it can reduce the computational complexity (\ie, FLOPs) of a wide range of transformer-like segmentation networks with a tiny drop of accuracy. Table~\ref{tab:citys} summarizes the performance and computation comparison between our framework and various state-of-the-art segmentation models. We mainly highlight the mIoU drop and FLOPs reduction rate in the brackets. With our method, the FLOPs of networks are reduced by 40\% $\sim$ 60\%, with less than 0.8\% mIoU loss. Especially, the mIoU is improved a little rather than reduced for some networks with the DoViT-S backbone, benefiting from the dynamic and sparse token pass. Moreover, we extend the DynamicViT~\cite{rao2021dynamicvit} backbone to segmentation architectures, abbreviated as ``DyViT/$\rho$'' ($0 \leq \rho \leq 1$ is the token ratio) in the table. The implicit token exiting strategy with regularization of keeping ratio, is insufficient for semantic segmentation models, in which the complex context confuses the token selection. As we can see that when the FLOPs are reduced by less than 20\%, the SETR drops more than 5\% mIoU. We also present the incremental parameters of DoViT and DyViT, relative to the standard ViT. To align with the embedding dimensions of larger transformers, the auxiliary heads with larger input dimensions introduce more extra parameters. But the extra parameters can be negligible comparing to the backbone itself. 

\begin{figure}[b]
   \centering
   \subfloat[Cityscapes]{
      \includegraphics[width=0.47\linewidth]{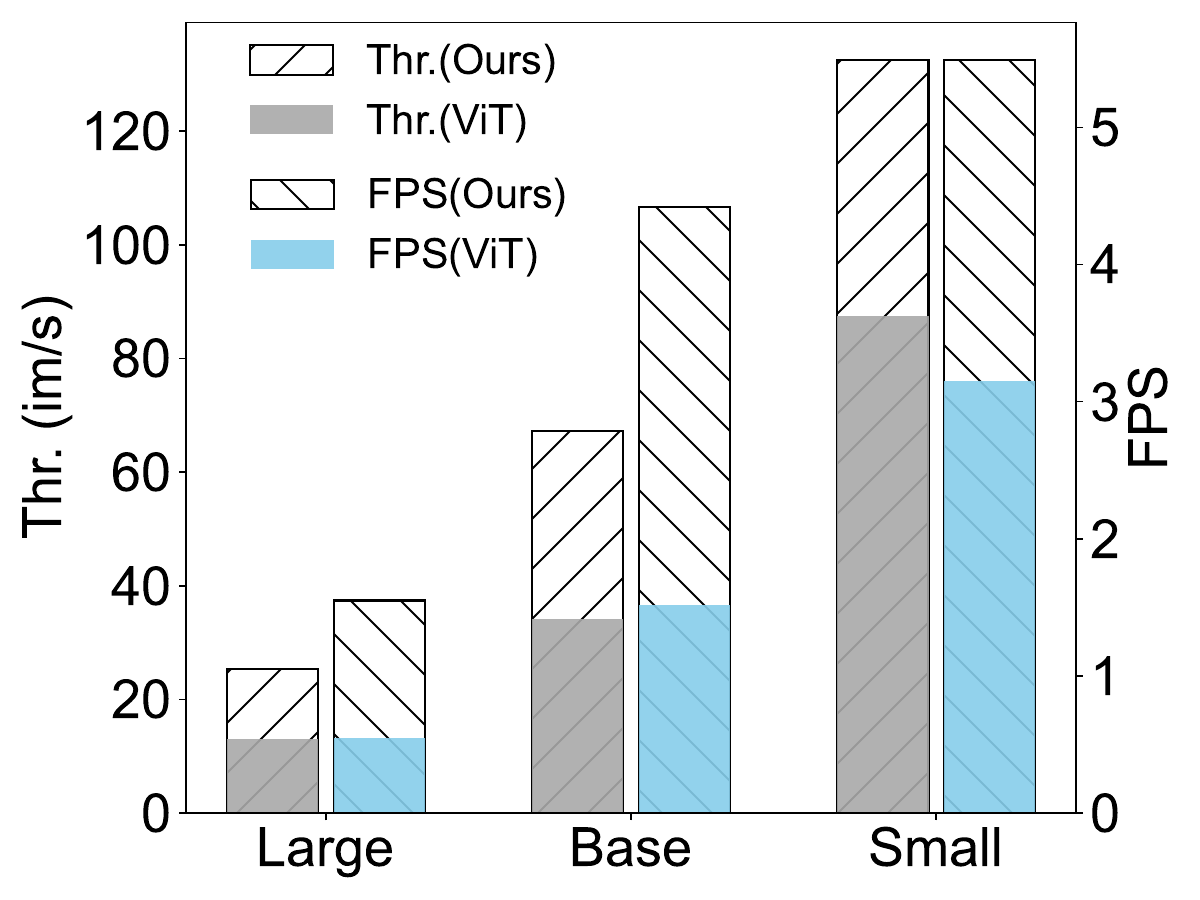}
      \label{fig:speedup_a}
   }
   \subfloat[ADE20K]{
      \includegraphics[width=0.47\linewidth]{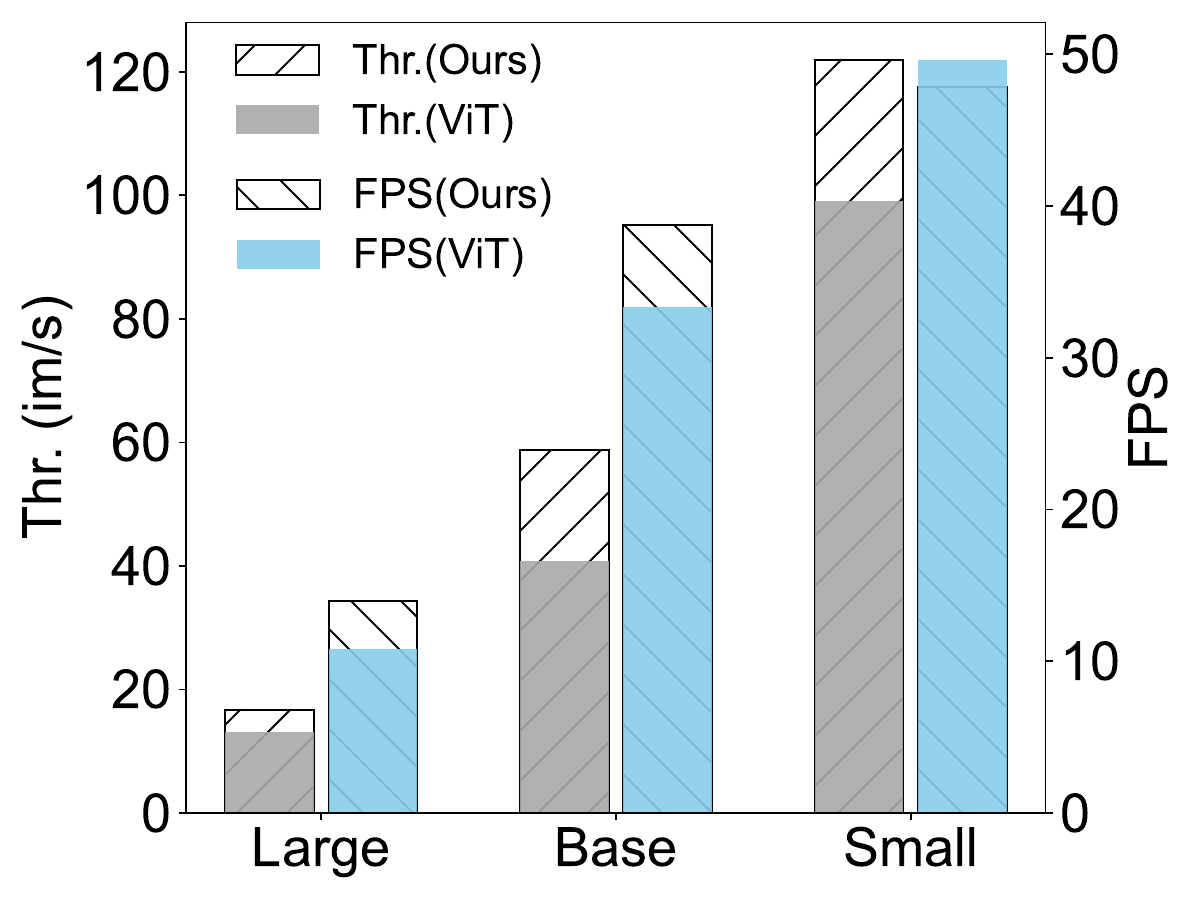}
      \label{fig:speedup_b}
   }
   \caption{The throughput (Thr.) and FPS improvement of SETR-DoViT-B on Cityscapes}
   \label{fig:speedup}
\end{figure}

\textbf{ADE20K.} 
To evaluate the effectiveness and efficiency of our approach, we conduct extensive experiments on the ADE20K dataset, as shown in Table~\ref{tab:ade}. We adopt various confidence threshold $\xi\in [0.96, 0.985]$ and compare the trade-offs between the mIoU performance and computational reduction. For SETR with ViT-base, our method can reduce 30\% FLOPs without mIoU drop. The ADE20K dataset is very challenging due to the large-scale semantic categories and complex scenes, making the models predict with lower confidence for numerous pixels. Even though leveraging a lower threshold can early stop more tokens and reduce more inference cost, it's difficult to reduce the FLOPs up to 20\%, especially for the small networks, \eg, DoViT-S, DoDeiT-S. Generally, the smaller networks achieve a less ratio of FLOPs reduction, caused by the lower confidence at early-probe. 

\textbf{Acceleration Effort.}
In Figure~\ref{fig:speedup}, we compare speedup on one NVIDIA V100 GPU in terms of SETR (ViT-B/DoViT-B) on two datasets respectively. It's worth noting that limited to the super-resolution, \ie, $1024\times 2048$ pixels, we utilize $512\times 512$ randomly cropped inputs to evaluate the throughput on Cityscapes. Figure~\ref{fig:speedup_a} illustrates that both the throughput and FPS of the large and base models are improved by over 2$\times$, without requiring hardware/library modification. Meanwhile, our method improves the throughput and FPS of ViT-large ($\xi=0.98$) variants by 27\% and 30\% on ADE20K (Figure~\ref{fig:speedup_b}). It is a pity that the throughput of ViT-small can be improved by 23\% on ADE20K, but the FPS decreases a little due to the extra computation of auxiliary heads.

\begin{table}[!t]
   \centering
   \resizebox*{\columnwidth}{!}{
     \begin{tabular}{p{1.4cm}<{\centering}|p{1.5cm}<{\centering}|p{0.7cm}<{\centering}|l|p{2cm}}
     \hline
     Network & Backbone & $\xi$ & mIoU (\%) & FLOPs (G) $\downarrow$ \\
     \hline
     \multirow{8}{*}{SETR} & ViT-B & --    & 46.37 & 88.54 \\
           & DoViT-B & 0.985 & 46.54 (+0.17) & 66.06 (-25\%) \\
           & DoViT-B & 0.98  & 46.41 (+0.04) & 63.16 (-29\%) \\
           & DoViT-B & 0.96  & 45.74 (-0.63) & 61.07 (-31\%) \\
         \cline{2-5}          & ViT-S & --    & 42.81 & 22.82 \\
           & DoViT-S & 0.985 & 42.56 (-0.25) & 19.85 (-13\%) \\
           & DoViT-S & 0.98  & 42.33 (-0.47) & 19.53 (-14\%) \\
           & DoViT-S & 0.96  & 42.26 (-0.55) & 18.15 (-20\%) \\
     \hline
     \multirow{2}{*}{SETR} & DeiT-B & --    & 43.67 & 88.54 \\
           & DoDeiT-B & 0.96  & 43.24 (-0.43) & 60.18 (-32\%) \\
     \hline
     \multirow{2}{*}{Segmenter} & ViT-S & --    & 46.19 & 26.57 \\
           & DoViT-S & 0.96  & 45.84 (-0.35) & 21.75 (-18\%) \\
     \hline
     \end{tabular}
   }
   \caption{Main results on ADE20K. We apply our method on SETR and Segmenter with different backbones. The mIoU performance and FLOPs reduction are reported on the validation set, when utilizing different thresholds $\xi$. }
   \label{tab:ade}
\end{table}

\subsection{Ablation Study}

\begin{table}[htbp]
   \centering
     \begin{tabular}{ccc|c}
     \toprule
     DoViT & Token Merging & Self-Distillation & mIoU \\
     \midrule
     \checkmark   &       &       & 75.37 \\
     \checkmark   & \checkmark   &       & 75.66 \\
     \checkmark   & \checkmark   & \checkmark   & 76.40 \\
     \bottomrule
     \end{tabular}
   \caption{Effects of different components in our framework. We provide the results after removing the self-distillation and token merging in terms of SETR-DoViT-B on Cityscapes. }
   \label{tab:ab}
\end{table}

\textbf{Effects of different components. } 
To verify the effectiveness of each component in our framework, we conduct ablation analysis on Cityscapes with SETR-DoViT-B, and present the results in Table~\ref{tab:ab}. With the token merging strategy, the mIoU of the DoviT is improved by about 0.3\%. With the pixel-wise self-distillation, the gap between the DoViT and ViT segmentation networks can be reduced to 0.2\% mIoU. 

\begin{figure}[!t]
\centering
\subfloat[computation and performance]{
   \includegraphics[width=0.47\linewidth]{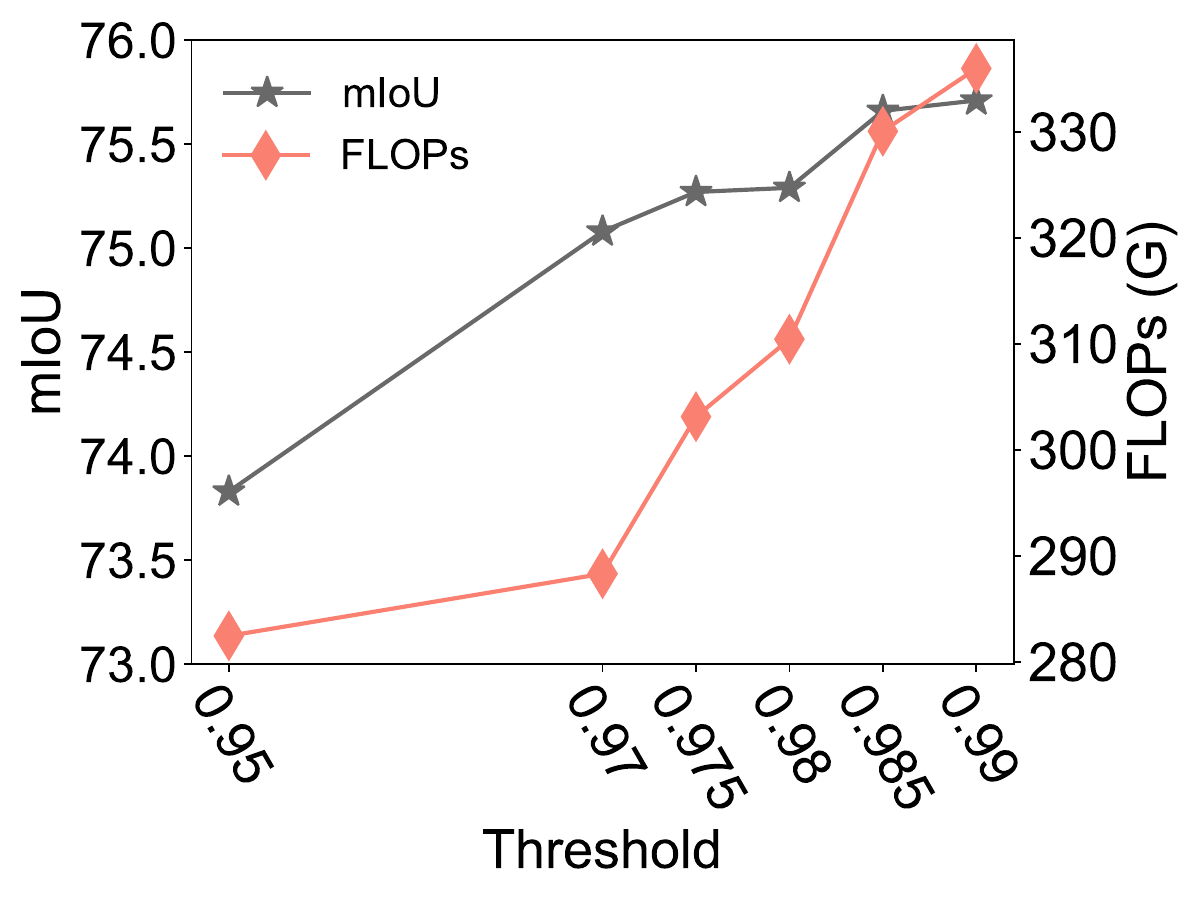}
   \label{fig:impact_a}
}
\subfloat[model acceleration]{
   \includegraphics[width=0.47\linewidth]{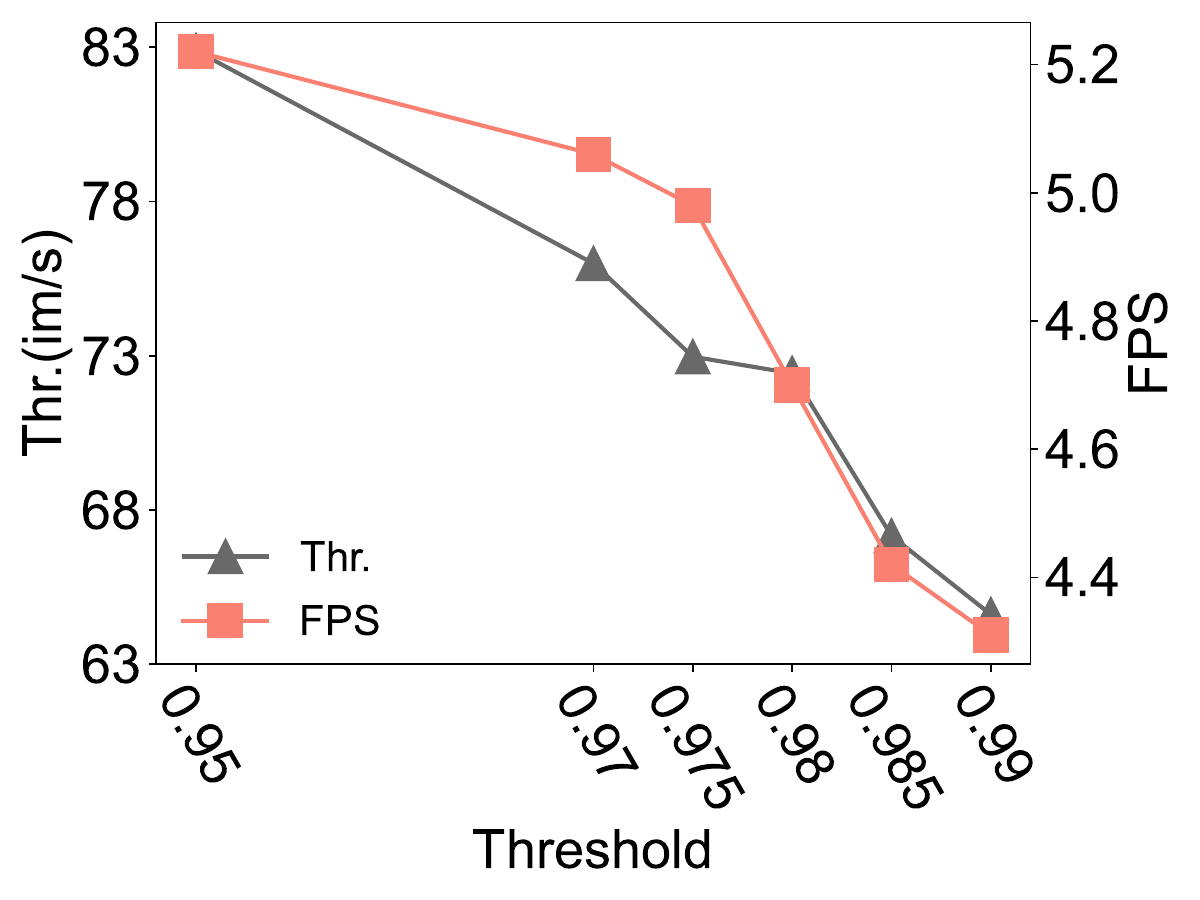}
   \label{fig:impact_b}
}
\caption{Impact of the confidence threshold $\xi$ to computation (a) and acceleration (b).}
\label{fig:impact}
\end{figure}

\textbf{Impact of the threshold. }
To investigate the impact of the confidence threshold $\xi$, we train SETR-DoViT-B on Cityscapes without self-distillation. Figure~\ref{fig:impact_a} depicts the trade-off between the performance and computation, varying the threshold from 0.95 to 0.99. It is obvious that with the increase of threshold, mIoU and FLOPs will increase, which is reasonable --- more computation brings better performance. In order to balance the computation and performance loss, it is suitable to set $\xi\in [0.985, 0.99]$ for Cityscapes. Thanks to the numerous easy-to-learn patches of cityscapes, the FLOPs can be reduced by 50\% when $\xi=0.99$. In addition, we plot the line of inference speed (throughput and FPS) with threshold, as shown in Figure~\ref{fig:impact_b}. Varying the threshold from 0.95 to 0.99, the throughput (Thr.) can be improved from 34.1 (as shown in Figure~\ref{fig:speedup_a}) to a range of $[63, 83]$, and the inference can be speeded up by at least $2.9\times$, \ie, from 1.52 FPS to over 4.42 FPS.

\begin{figure*}[!t]
   \centering
   \includegraphics[width=.96\linewidth]{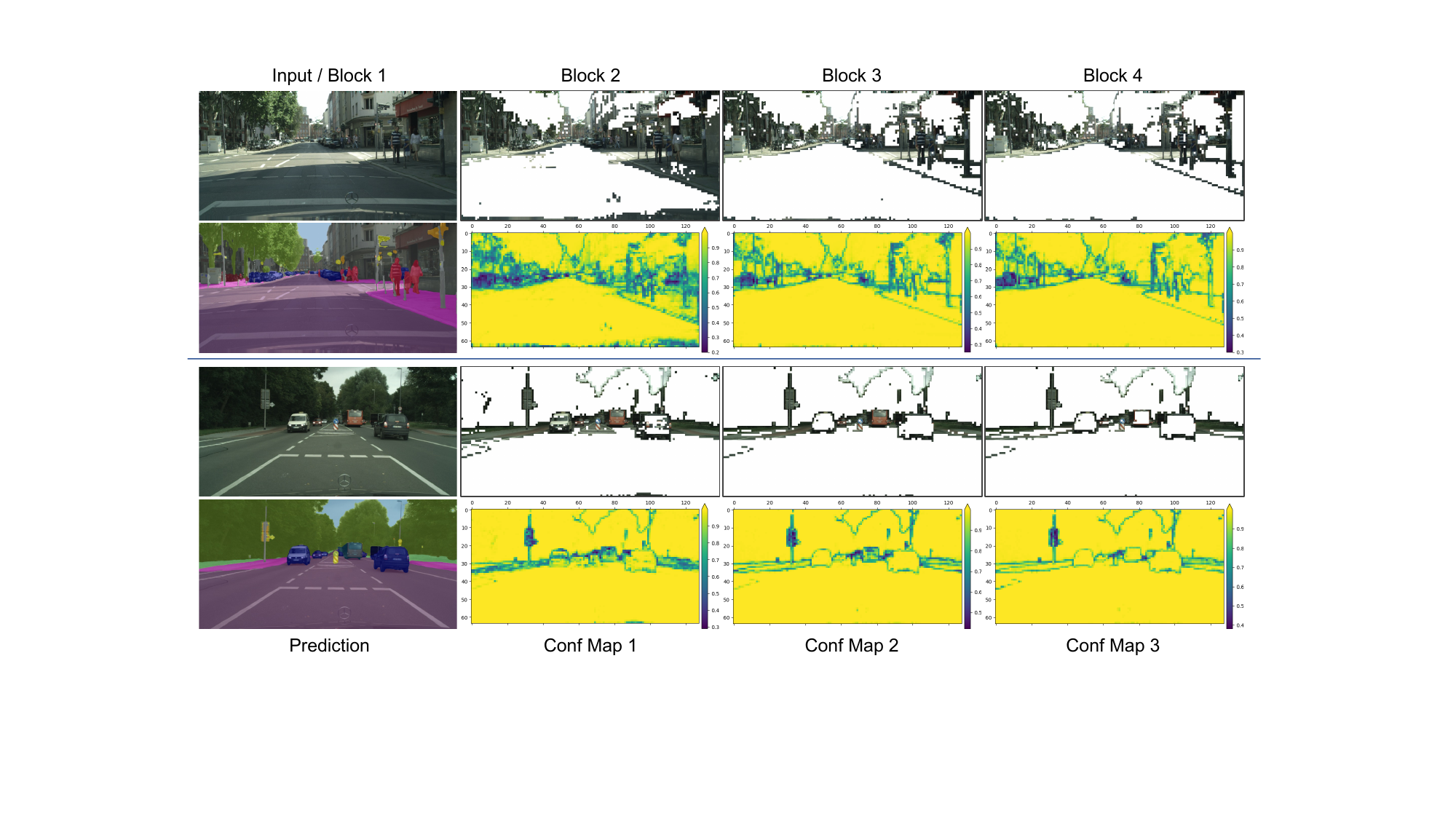}
   \caption{Visualizations of keeping tokens and segmentation results of two images from Cityscapes. The first and third rows depict the corresponding patches of the keeping tokens input to each block, where the white region is corresponding to the stopping tokens. The second and fourth rows show the final prediction results and confidence score map at each block. }
   \label{fig:vis}
\end{figure*}

\begin{figure}[t]
\centering
\subfloat[Case 1]{
   \includegraphics[width=0.47\linewidth]{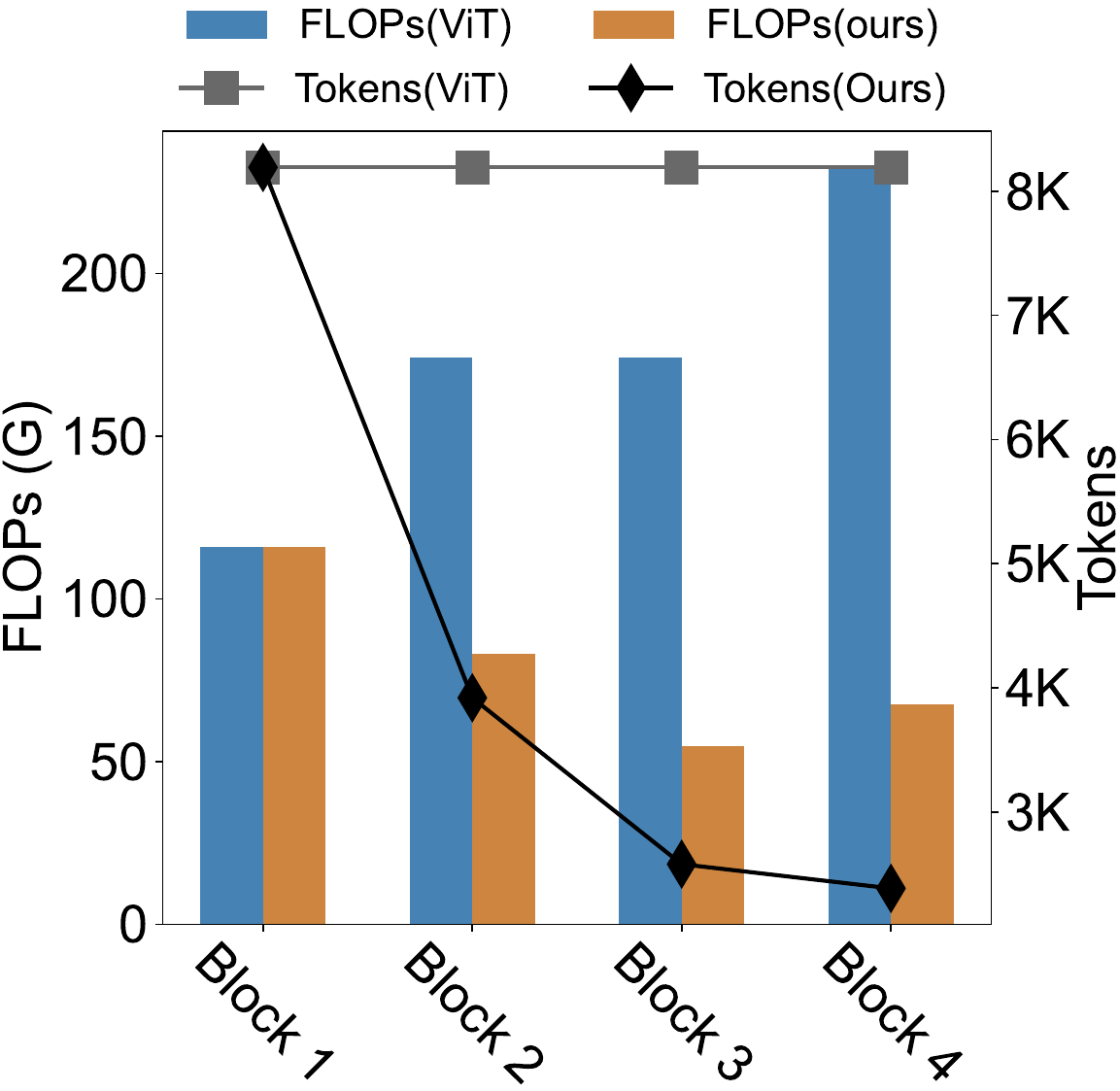}
}
\subfloat[Case 2]{
   \includegraphics[width=0.47\linewidth]{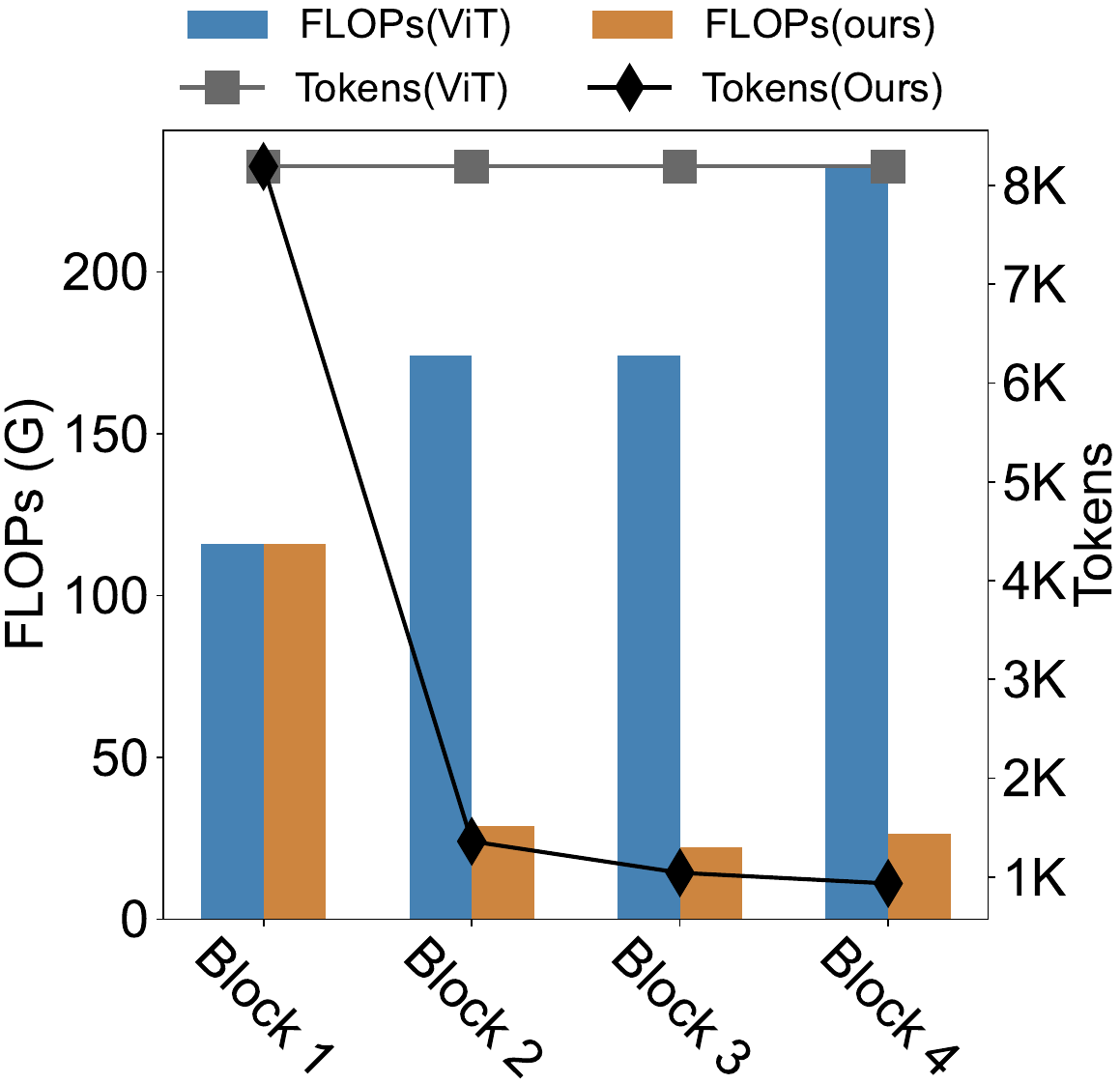}
}
\caption{Case study of two images in Cityscapes. We report the number of keeping tokens and FLOPs at each transformer block, with a $1024 \times 2048$ input to SETR with ViT-B/DoViT-B. The total number of tokens is 8,192 as the patch size is set to 16. }
\label{fig:case}
\end{figure}

\subsection{Visualization}

In Figure~\ref{fig:vis}, we show the qualitative results of two cityscape images. We find that applying DoViT results in progressive reduction of keeping tokens/patches when forwarding block-by-block. Meanwhile, the easy-to-learn patches, such as that cover road, sky and tree, are stopped from self-attention early, while the hard patches consisting of complex context, will be kept until the end of the vision transformers. Additionally, we visualize the confidence score maps predited by the three auxiliary heads. The tokens/patches with higher scores (closer to yellow) are supposed to be removed from the calculation. What's more interesting is that, though the tokens covering easy patches are removed, some more informative edge parts are preserved, \eg, the outline of the trees and cars. 
Figure~\ref{fig:case} demonstrates the corresponding quantitative information of the two inference cases. The number of tokens involving in the four ViT blocks are all 8,192. In case 1 (a), over half of the tokens are removed at the second block, and the FLOPs per block also drops accordingly. In case 2 (b), over 80\% tokens are stopped at the second block, and only 10\% tokens are kept at the last block. These results reflect the efficiency and interpretability of our dynamic token-pass method. 
The early-probe scheme determines token-pass adaptively, rather than forcely stopping a fixed ratio of tokens for all images. Thus, the inference costs of different images could be very different.



\section{Conclusion}

In this work, we explore the segmentation transformer acceleration from a perspective of data-redundancy. We have introduced dynamic token-pass transformers (DoViT) to adaptively adjust the inference cost based on input complexity. DoViT gradually reduces the number of tokens passing self-attention layer and shorts the hierarchical-stopped tokens into a unified decoder. We evaluate the effectiveness of our approach in computation reduction and inference speedup, and discuss some meaningful issues. 
In the future, we plan to combine our data-aware transformer acceleration method with the parameter-aware model compression approaches and extend it to other dense prediction tasks.

{\small
\bibliographystyle{ieee_fullname}
\bibliography{ref}
}

\end{document}